\documentclass[11pt]{article}

\usepackage[final]{acl}

\usepackage{times}
\usepackage{latexsym}

\usepackage[T1]{fontenc}

\usepackage[utf8]{inputenc}

\usepackage{microtype}

\usepackage{inconsolata}

\usepackage{graphicx}

\usepackage{amsmath, amssymb}
\usepackage{cleveref}
\usepackage{hyperref}
\usepackage[most,skins]{tcolorbox}
\usepackage{enumitem}
\usepackage{xcolor}
\tcbset{
  promptbox/.style={
    width=\linewidth,
    top=8pt,
    bottom=2pt,
    left=4pt,
    right=4pt,
    colback=gray!1!white,
    colframe=black,
    colbacktitle=black,
    enhanced,
    breakable,
    center,
    fontupper=\normalsize,
    before upper={\setlength{\baselineskip}{1.12\baselineskip}}, 
    attach boxed title to top left={yshift=-0.1in,xshift=0.15in},
    boxed title style={boxrule=0pt,colframe=white,},
  }
}
\newtcolorbox{promptbox}[2][]{promptbox,title=#2,#1}

\usepackage{orcidlink}

\usepackage[table]{xcolor}
\newcommand{\oursbg}{\cellcolor{gray!10}}

\newcommand{\tablestyle}[2]{\setlength{\tabcolsep}{#1}\renewcommand{\arraystretch}{#2}\centering\footnotesize}
\usepackage{makecell}
\usepackage{multicol}
\usepackage{multirow}
\usepackage{xcolor} 
\usepackage{colortbl} 
\definecolor{gr}{gray}{0.95} 
\definecolor{convcolor}{HTML}{412F8A}   
\definecolor{vitcolor}{HTML}{fc8e62}    

\newcommand{\convcolor}[1]{\textcolor{convcolor}{#1}}
\newcommand{\vitcolor}[1]{\textcolor{vitcolor}{#1}}

\newcommand{\vb}{\vitcolor{$\mathbf{\circ}$\,}}  
\newcommand{\cb}{\convcolor{$\bullet$\,}}        

\usepackage{pifont}
\newcommand{\cmark}{\ding{51}} 
\newcommand{\xmark}{\ding{55}} 

\title{Re$^3$Cap: Retrieval-Guided Refinement for Image Captioning Enhancement via Reinforcement Learning}

\author{
Haonan Jia\textsuperscript{1}\thanks{Equal contribution.},
Shichao Dong\textsuperscript{1}\footnotemark[1],
Zenghui Sun\textsuperscript{1},
Jiawen Zheng\textsuperscript{2},
Ziqi Miao\textsuperscript{3}, \\ \bfseries
Gege Shi\textsuperscript{1},
Qiuyu Zhao\textsuperscript{1},
Jinsong Lan\textsuperscript{1},
Xiaoyong Zhu\textsuperscript{1},
Bo Zheng\textsuperscript{1}\thanks{Corresponding author.}\\
\textsuperscript{1}Taobao \& Tmall Group of Alibaba \\
\textsuperscript{2}The Hong Kong University of Science and Technology (Guangzhou)\\
\textsuperscript{3}Shanghai Artificial Intelligence Laboratory\\
}

\begin{document}
\maketitle

\begin{abstract}

Reinforcement Learning (RL) has demonstrated significant gains in image captioning, yet it is still limited in encouraging Large Vision-Language Models (LVLMs) to explore novel reasoning strategies. 
This limitation leads to a performance gap between RL and Supervised Fine-Tuning (SFT).
In this paper, we argue that multi-modal retrieval can serve as an effective reasoning signal for caption refinement.
Based on this insight, we present the Retrieval-Guided Refinement for Image Captioning (Re$^3$Cap), a retrieval-guided reasoning strategy that enhances image captioning without requiring additional annotations.
Instantiated by Caption Refinement Suggester (CRS) and Caption Quality Assessor (CQA), this strategy identifies hallucinations and omissions in image captions, leading to more accurate and detailed descriptions.
Extensive experiments demonstrate the superiority of our method in image captioning, even compared with Supervised Fine-Tuning. 
Especially, Re$^3$Cap outperforms GRPO with an average improvement of 8.64\% in relation reasoning on the COCO-LN500 benchmark. 
The code will be released when the paper is accepted.

\end{abstract} 
\section{Introduction}
\label{sec:intro}
Image captioning~\cite{karpathy2015deep,huang2019attention,liu2017improved} is a fundamental task in computer vision and plays an essential role in various applications, such as text-image retrieval~\cite{duan2025fuzzy,chen2024make}, text-to-image generation~\cite{betker2023improving, zheng2024cogview3}, and visual question answering~\cite{cheng2025simplevqa,hu2024omnimedvqa,miao2026seeing}.
Recently, the development of Large Vision-Language Models (LVLMs)~\cite{instructblip,liu2023improvedllava,dong2024internlm,bai2023qwen,ye2023mplugowl} has demonstrated notable success in multi-modal understanding and yielded significant performance gains in image captioning.
Nevertheless, image captions generated by existing methods ~\cite{cornia2020meshed, huang2019attention, liu2017attention, liu2017improved,feng2019unsupervised, bahng2025cycle,tewel2022zerocap, xu2023zero} are prone to hallucinations and often fail to capture fine-grained visual details. 
Consequently, it remains challenging to generate detailed and accurate image captions.

Previous studies have primarily leveraged reinforcement learning (RL) to post-train Large Vision-Language Models (LVLMs) to enhance their image captioning capabilities.
For instance, CLIP-based methods~\cite{cho2022fine, yu2023cgt, dzabraev2024vlrm} assess the correlation score between images and LVLM-generated captions based on Vision Language Models (VLMs).
By using this score as the reward signal, these methods force LVLMs to generate more detailed image captions.
Unfortunately, due to the constrained compositional reasoning capabilities of VLMs ~\cite{wang2024diagnosing}, these methods remain highly susceptible to reward hacking.
Accordingly, SC-Captioner~\cite{zhang2025sc} annotates keywords for each image and evaluates the quality of image captions by checking whether the caption explicitly contains these words.
However, RL-based approaches still lag behind Supervised Fine-Tuning methods ~\cite{luo2024unleashing, yang2025bacon, li2024densefusion, chen2024sharegpt4v}.

Recent studies~\cite{yue2025does} reveal that reinforcement learning merely selects the highest-reward caption from candidates that are pre-generated by LVLMs. 
During training, LVLMs exhibit limited exploration of novel reasoning strategies, failing to produce diverse and previously unexplored caption candidates.
In this case, the captioning performance of RL-optimized LVLMs remains bounded by the intrinsic reasoning capabilities of the base model.
Consequently, the crucial challenge in image captioning lies in exploring novel reasoning strategies that empower models to generate unexplored candidate captions for RL.

In this paper, we argue that multi-modal retrieval can serve as an effective reasoning signal for caption refinement.
Intuitively, \textbf{visually similar images often share overlapping semantic content}.
In this way, by using the source image as a query, we can infer its semantic content from descriptions of the retrieved similar images.
Moreover, \textbf{semantically similar queries tend to yield consistent retrieval results}.
Consequently, a detailed and accurate image caption should induce retrieval results similar to those obtained from the source image. 
Any discrepancy between image-based and caption-based retrieval descriptions signals the misalignment between the image and the LVLMs-generated caption, indicating hallucinations.


Building on these observations, we introduce Re$^3$Cap, a retrieval-guided reasoning strategy that enhances image captioning through two components: the Caption Refinement Suggester (CRS) and the Caption Quality Assessor (CQA).
Specifically, CRS identifies critical elements to preserve in the image caption by verifying descriptions that consistently overlap across retrieved similar images.
Furthermore, CQA analyzes discrepancies between image-based and caption-based retrieval descriptions to indicate hallucinations and omissions in the generated caption.
Through this process, we determine which elements in LVLM-generated captions should be retained, which hallucinated content needs to be removed, and which visual details from the source image have been omitted.
During RL training, such reasoning results will guide LVLMs to refine their captions without requiring additional annotations.
By injecting this reasoning strategy, LVLMs generate diverse, previously unexplored caption candidates, thereby significantly enhancing their image captioning capability.
Extensive experiments demonstrate the effectiveness of our proposed method across multiple LVLM architectures under diverse reward functions.
Our contributions can be summarized as follows:
\begin{itemize}
    \item We present a novel retrieval-based reasoning strategy to indicate hallucinations and omissions in captions without requiring additional annotations.
    \item We propose Re$^3$Cap, which guides LVLMs to generate previously unexplored caption candidates, thereby enhancing model performance in image captioning.
    \item Extensive experiments demonstrate that our method improved the performance on various LVLMs, outperforming state-of-the-art methods by a large margin, even compared with Supervised Fine-Tuning.
\end{itemize}
\section{Related Work}
Image captioning is a fundamental task in computer vision, serving as a key bridge between the visual and linguistic modalities.
Recent works can be broadly categorized into two lines: supervised fine-tuning (SFT) and reinforcement learning (RL).

\subsection{Supervised Fine-Tuning}
Previous approaches typically adopt an encoder–decoder paradigm, where an encoder extracts visual representations from the image, and a decoder autoregressively generates the caption~\cite{cornia2020meshed, huang2019attention, liu2017attention, liu2017improved, vinyals2015show, wang2022git, mokady2021clipcap, luo2023tuning}.
Building upon the encoder–decoder paradigm, several works further incorporate retrieval-augmented generation (RAG), where the captioner is conditioned not only on the image but also on relevant texts retrieved from external corpora~\cite{ramos2023smallcap, li2024evcap, kim2025vipcap}.
Complementary approaches substitute human annotations with synthetic data for supervision~\cite{luo2024unleashing, yang2025bacon, li2024densefusion, chen2024sharegpt4v}.
In addition, some approaches perform self-supervised training by leveraging the shared multimodal embedding space of vision–language models~\cite{fei2023transferable, tam2023simple, lee2025diffusion}.
Controllability has also been explored by fine-tuning captioning models to obey user-specified control signals~\cite{kornblith2023guiding, saito2025captionsmiths}.
Despite substantial gains in caption accuracy and detail, these methods still heavily depend on large-scale image–caption datasets, which are expensive and time-consuming to collect.

\begin{figure*}[t]
  \centering
    \includegraphics[width=\linewidth]{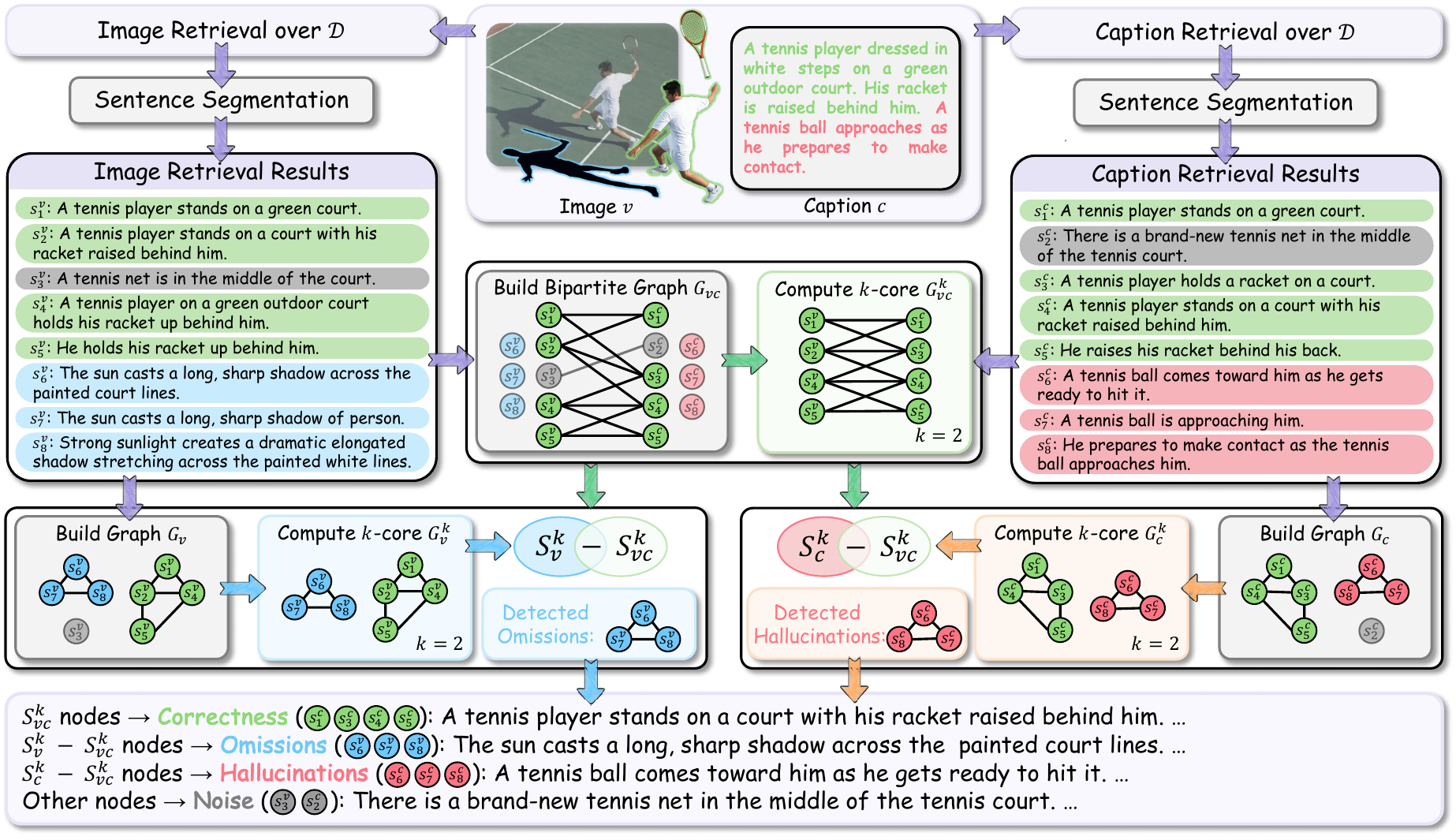}
  \caption{\textbf{Reasoning strategy of Caption Refinement Suggester (CRS) and Caption Quality Assessor (CQA).} We leverage k-core subgraph computation to analyze critical textual content in the retrieval results. Specifically, CRS extracts overlapping sentences $S^k_{v}$ from the image retrieval results and uses them as signals to guide LVLMs to incorporate these key elements into their generated descriptions. By analyzing the discrepancy between $S^k_{v}$ and caption-retrieved results $S^k_{c}$, CQA identifies hallucinations ($S^k_c-S^k_{vc}$) and omissions ($S^k_v-S^k_{vc}$) in this caption.}
  \label{fig:k-core}
\end{figure*}

\subsection{Reinforcement Learning}
Increasingly, researchers adopt reinforcement learning to improve image captioning in LVLMs by optimizing task-specific reward signals.
For instance, CLIP-based methods~\cite{cho2022fine, yu2023cgt, dzabraev2024vlrm} assess the correlation score between images and LVLM-generated captions based on Vision Language Models (VLMs)~\cite{radford2021learning}.
Some approaches impose cycle-consistency by regenerating the image from the caption and using the reconstruction fidelity as the training signal~\cite{feng2019unsupervised, bahng2025cycle}. 
Additionally, some methods use a self-retrieval objective, encouraging captions that can successfully retrieve originating images~\cite{liu2018show, gaur2024no, dessi2023cross}.
Furthermore, reinforcement learning has been used to promote self-correction in captioning models~\cite{zhang2025sc}.
Additionally, some work uses caption-conditioned downstream VQA accuracy as a reward signal~\cite{xing2025caprl}.
More recent work boosts the precision and detail richness of captions by minimizing information loss in modality conversion~\cite{jia2026cim}.
However, RL-based methods still lag behind SFT.


\section{Method}
In this section, we introduce Retrieval-Guided Refinement for Image Captioning (Re$^3$Cap), a reasoning strategy to enhance the image captioning of LVLMs.
Specifically, Caption Refinement Suggester (CRS) first identifies critical semantic elements within image captions. 
Subsequently, the Caption Quality Assessor (CQA) identifies omissions or misrepresentations in image captions.
Leveraging the above guidance, our method finally encourages LVLMs to generate previously unexplored caption candidates. 

\subsection{Caption Refinement Suggester}
\textit{Visually similar images often share overlapping semantic content.}
Based on this insight, the Caption Refinement Suggester (CRS) analyzes semantic elements that consistently appear across visually similar images. 
It then suggests LVLMs to incorporate these elements to refine their generated captions.

As shown in \Cref{fig:k-core}, let $v$ denote the image.
The dataset $\mathcal{D} = \{p_{i}\}_{i=1}^N$ consists of $N$ image-text pairs $p_{i}$, each containing an image $x_i$ and its corresponding text $t_i$.
With $v$ as query, we perform image retrieval over the dataset $\mathcal{D}$ to obtain the top-K retrieval results denoted as $\mathcal{R}_v = \operatorname{TopK}(\{\mathrm{SIM}(v, x_{i})\}_{i=1}^{N}\bigr) = \{p^v_i\}_{i=1}^K$.
$\mathrm{SIM}(v, x_i)$ means the correlation score between $v$ and $x_{i}$ calculated by the image retrieval model~\cite{cherti2023reproducible}.
Through this process, we obtain a set of image-text pairs $R_v$.
Each image in $R_v$ shares similar visual representations to the query image $v$.
Moreover, we construct a graph $G_{v}$ to model descriptions corresponding to images in $R_v$. 
In $G_v$, each sentence is treated as a node.
The textual similarity score between every pair of nodes is computed by SBERT~\cite{reimers2019sentence}. 
When the similarity score exceeds a predefined threshold $\tau$, we establish an edge between these two nodes.
Following algorithm~\cite{seidman1983network}, we compute its $k$-core subgraph $G^k_{v}=(S^{k}_{v},E^{k}_{v})$ to analyze semantically consistent elements in $G_{v}$.
$E^k_v$ and $S^k_v$ denote edges and nodes in the graph.
By decomposing the $k$-core, CRS filters out long-tail descriptions in $\mathcal{R}_v$ and retains semantically consistent elements across image retrieval results. 

In this way, our method leverages $k$-core analysis in visually similar images to identify semantic content (i.e., $S^k_v$) corresponding to the query image. 
During caption refinement, CRS suggests LVLMs to incorporate these descriptions, thereby improving the accuracy of the refined image caption.

\subsection{Caption Quality Assessor}
\textit{Semantically similar queries tend to yield consistent retrieval results.}
Motivated by this observation, the Caption Quality Assessor (CQA) evaluates discrepancies between the query image and its caption by comparing their respective retrieval results.
By prompting LVLMs with the identified discrepancies between the image and its description, CQA guides LVLMs to generate more accurate captions.

Let $c$ be the caption generated by the LVLM for the query image $v$, as illustrated in ~\Cref{fig:k-core}.
We then perform text retrieval over the dataset $\mathcal{D}$, using $c$ as the query, to obtain the top-$K$ retrieval results: $\mathcal{R}_c = \operatorname{TopK}(\{\mathrm{SIM}(c, t_i)\}_{i=1}^{N}) = \{p^c_i\}_{i=1}^{K}$.
Similar to CRS, we construct a graph $G_{c}$ over sentences in the caption-retrieved results $\mathcal{R}_c$ and compute its $k$-core subgraph $G^k_{c}=(S^{k}_{c},E^{k}_{c})$. 
Each node in $G^k_{c}$ corresponds to a sentence that appears densely in caption retrieval results. 
We further construct a bipartite graph over $\mathcal{R}_v \cup \mathcal{R}_c$ and compute its $k$-core subgraph $G^k_{vc}=(S^k_{vc},E^k_{vc})$, which captures the semantically consistent content shared by the image and its corresponding caption.
In this way, we can characterize the discrepancy between the image and its caption from two perspectives: hallucinated content in the caption is measured as $S^k_c - S^k_{vc}$, while $S^k_v - S^k_{vc}$ represents critical semantic content omitted by the LVLM in its generated caption.


Through information retrieval, CQA reformulates the complex cross-modal task of image caption quality assessment as an analysis of textual discrepancies within the retrieved results.
As a result, the module can identify hallucinations and omissions in image captions without requiring additional annotations.

\begin{figure*}[t]
  \centering
    \includegraphics[width=0.8\linewidth]{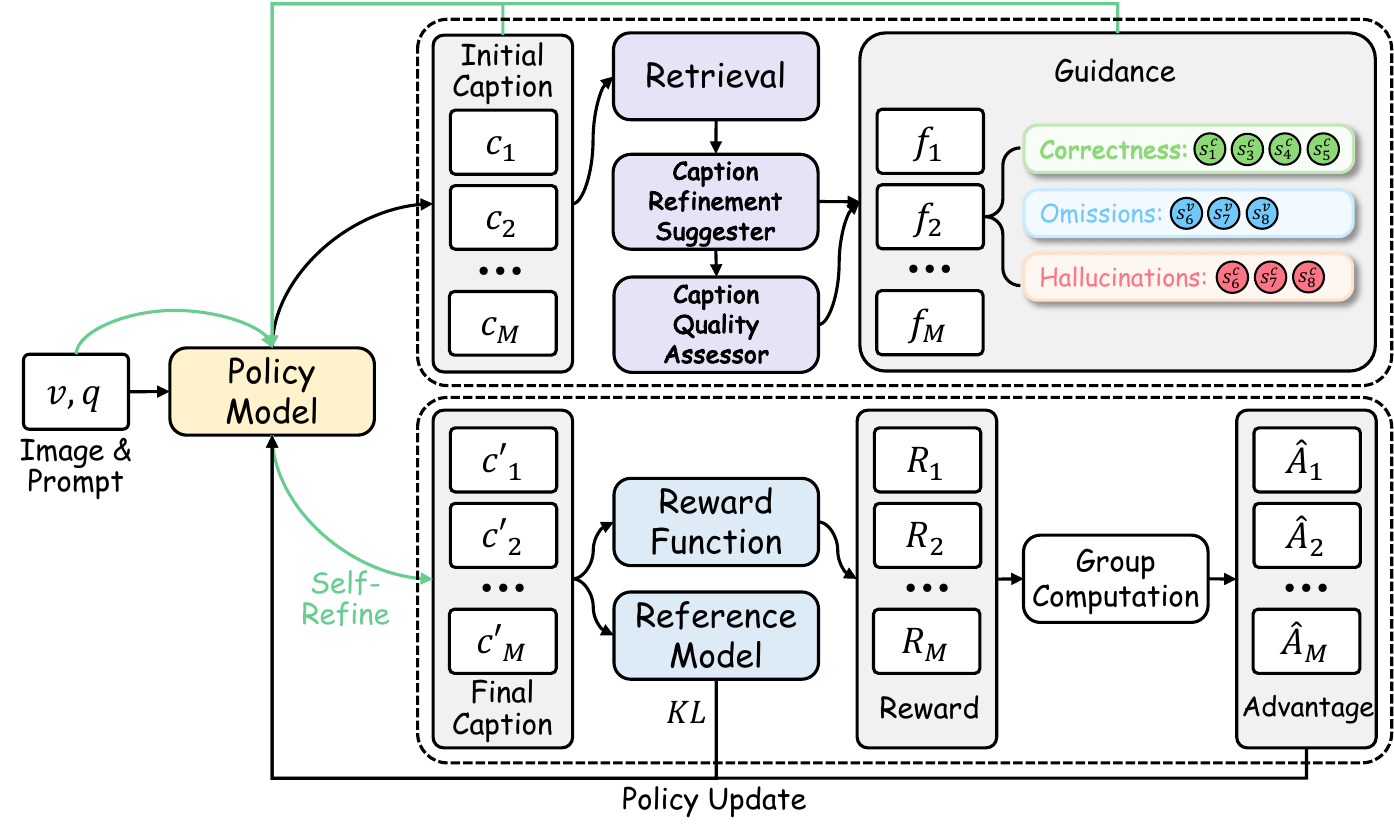}
  \caption{\textbf{Overview of Retrieval-Guided Refinement for Image Captioning (Re$^3$Cap).} Re$^3$Cap begins by sampling initial captions $\{c_i\}_{i=1}^{M}$ for each image $v$. Next, it performs image-conditioned retrieval and caption-conditioned retrieval, and leverages k-core analysis to generate guidance $\{f_i\}_{i=1}^{M}$ based on Caption Refinement Suggester and Caption Quality Assessor. Finally, the method incorporates the guidance into prompts to obtain refined captions $\{c’_i\}_{i=1}^{M}$ and optimizes the policy model with reinforcement learning.}
  \label{fig:framework}
\end{figure*}

\subsection{Overview of Re$^3$Cap}
Based on the above reasoning strategy, we present the Retrieval-Guided Refinement for Image Captioning (Re$^3$Cap), a reinforcement learning framework that enhances image captioning without requiring additional annotations.
As shown in \Cref{fig:framework}, given an input image $v$ and a prompt $q$, we firstly sample a group of initial captions $\{c_i\}_{i=1}^{M}$ using LVLMs.
With input image and initial captions as the queries, we perform image-conditioned retrieval and caption-conditioned retrieval.
In this way, we leverage k-core analysis over retrieval results to generate guidance for caption improvement based on the Caption Refinement Suggester and the Caption Quality Assessor.
The guidance $\{f_i\}_{i=1}^{M}$ identifies the correct elements, hallucinations, and omitted semantic content in LVLM-generated captions.
Using the guidance, we further prompt the LVLMs to produce refined captions $\{c’_i\}_{i=1}^{M}$ that are more accurate and detailed.
The reward function scores each refined caption with a reward $R_i$.


To enable the model to generate higher-quality captions directly from images during inference, Re$^3$Cap removes the initial caption and guidance used in the rollout stage from the policy input during policy optimization.
This creates an off-policy optimization setting: the sampled captions are generated by a behavior policy conditioned on the image, the initial caption, and the guidance, whereas the optimized policy is conditioned only on the image.
To correct for this distribution mismatch while still preserving a trust-region center for regularizing the policy update, we decouple the proximal policy from the behavior policy, following the decoupled PPO formulation~\cite{hilton2022batch, fu2026areal}.
Specifically, we optimize the following objective:
\[
\begin{aligned}
\mathcal{J}(\theta)
\!&=\!\mathbb{E}_{v\sim\mathcal{V},\{c_i\}_{i=1}^{M}\sim\pi_{\theta_{\mathrm{old}}}(\cdot\mid v),\{c'_i\}_{i=1}^{M}\sim\pi_{\theta_{\mathrm{old}}}(\cdot\mid v,c_i,f_i)} \\
&\frac{1}{M}\sum_{i=1}^{M}\frac{1}{|c'_i|}\sum_{t=1}^{|c'_i|}
(\min(\frac{\pi_{\theta}}{\pi_{\mathrm{behav}}}\hat{A}_{i,t}, \\
&\frac{\pi_{\mathrm{prox}}}{\pi_{\mathrm{behav}}}\operatorname{clip}(\frac{\pi_{\theta}}{\pi_{\mathrm{prox}}},1-\epsilon,1+\epsilon)\hat{A}_{i,t}) \\
&-\beta D_{\mathrm{KL}}(\pi_{\theta}\|\pi_{\mathrm{ref}})),
\end{aligned}
\]
where
\[
\begin{aligned}
\pi_{\mathrm{behav}}
&=
\pi_{\theta_{\mathrm{old}}}
\left(c'_{i,t}\mid v,q,c_i,f_i,c'_{i,<t}\right), \\
\pi_{\mathrm{prox}}
&=
\pi_{\theta_{\mathrm{old}}}
\left(c'_{i,t}\mid v,q,c'_{i,<t}\right), \\
\pi_{\theta}
&=
\pi_{\theta}
\left(c'_{i,t}\mid v,q,c'_{i,<t}\right).
\end{aligned}
\]
As a result, the optimized policy is conditioned solely on the image, eliminating the need for retrieval or k-core analysis at inference time.
\section{Experiment}
In this section, we first introduce our experimental settings. 
We then present a reasoning capability analysis.
Subsequently, we demonstrate the effectiveness of our method by comparing it with GRPO and state-of-the-art image captioning methods.
Finally, we present an ablation study to investigate the contribution of each component.
Additional experiments, including robustness across diverse encoders, sensitivity to hyperparameter choices such as the retrieval number $K$ and the similarity threshold $\tau$, and computational overhead, are provided in the \Cref{sec:b}.

\subsection{Experimental Settings}
\textbf{Training settings.}
Following ~\cite{zhang2025sc} and CIM~\cite{jia2026cim}, we use images from the RefinedCaps dataset~\cite{zhang2025sc} as the training set, consisting of 6.5K images sampled from the COCO training split~\cite{lin2014microsoft}.
For both image-conditioned and caption-conditioned retrieval, we retrieve the top-K candidates ($K = 3$) from a dataset constructed by augmenting RefinedCaps~\cite{zhang2025sc} with DenseFusion-1M~\cite{li2024densefusion}.
To avoid data leakage, we ensure that the retrieval corpus is disjoint from all evaluation benchmarks.
We use SBERT~\cite{reimers2019sentence} with MPNet-base backbone~\cite{song2020mpnet} as the text encoder and OpenCLIP ViT-H/14~\cite{cherti2023reproducible} as the image encoder, respectively.
For CRS and CQA, we set the $k$ in the $k$-core to $k = \lceil K/2 \rceil = 2$, and use a threshold $\tau = 0.7$.
We adopt the VERL framework~\cite{sheng2025hybridflow} for training. 
For hyperparameters, we utilize the Adam optimizer and train for two epochs with a constant learning rate of $1\times10^{-6}$.
For rollout, the prompt batch size is 256, and we sample $M = 5$ responses for each prompt.
For training, the mini-batch size is set to 64. 
We set the clipping ratio to $\epsilon = 0.2$ and the KL penalty coefficient to $\beta = 0.001$.

\noindent\textbf{Models.}
To validate the generalizability of Re$^3$Cap, we evaluate its performance across representative Large Vision-Language Models (LVLMs): LLaVA-1.5-7B~\cite{liu2023improvedllava}, Qwen2-VL-7B~\cite{wang2024qwen2}, and Qwen2.5-VL-7B~\cite{bai2025qwen2}.
Additional evaluations on InternVL3-8B~\cite{zhu2025internvl3} and Qwen3-VL-8B~\cite{bai2025qwen3} are provided in \Cref{sec:b.1}.

\noindent\textbf{Benchmarks.}
We use COCO-LN500~\cite{pont2020connecting} and DOCCI500~\cite{onoe2024docci} as the evaluation benchmarks to validate the effectiveness of our proposed Re$^3$Cap.
COCO-LN500~\cite{pont2020connecting} consists of 500 image–caption pairs from the Localized-narratives test set in COCO2017~\cite{lin2014microsoft}.
DOCCI500~\cite{onoe2024docci} is a random sample of 500 image-caption pairs from DOCCI test split, where images largely lack human-centric content.

\noindent\textbf{Metrics.}
Following SC-Captioner~\cite{zhang2025sc}, we evaluate caption quality using the F1 score from three aspects: objects, attributes, and relations.
For relations, we measure relational correctness via VQA-based accuracy based on Qwen3~\cite{yang2025qwen3}.

\noindent\textbf{Baselines.}
We adopt Group Relative Policy Optimization (GRPO)~\cite{shao2024deepseekmath} as the baseline with multiple reward functions. 
We consider CLIP~\cite{cho2022fine}, which uses the CLIP~\cite{radford2021learning} image-text similarity score as the reward; SC~\cite{zhang2025sc}, which rewards keyword-level self-correction; and CIM~\cite{jia2026cim}, which rewards the similarity between images retrieved by the caption and original image.

\begin{figure}[t]
  \centering
    \includegraphics[width=\columnwidth]{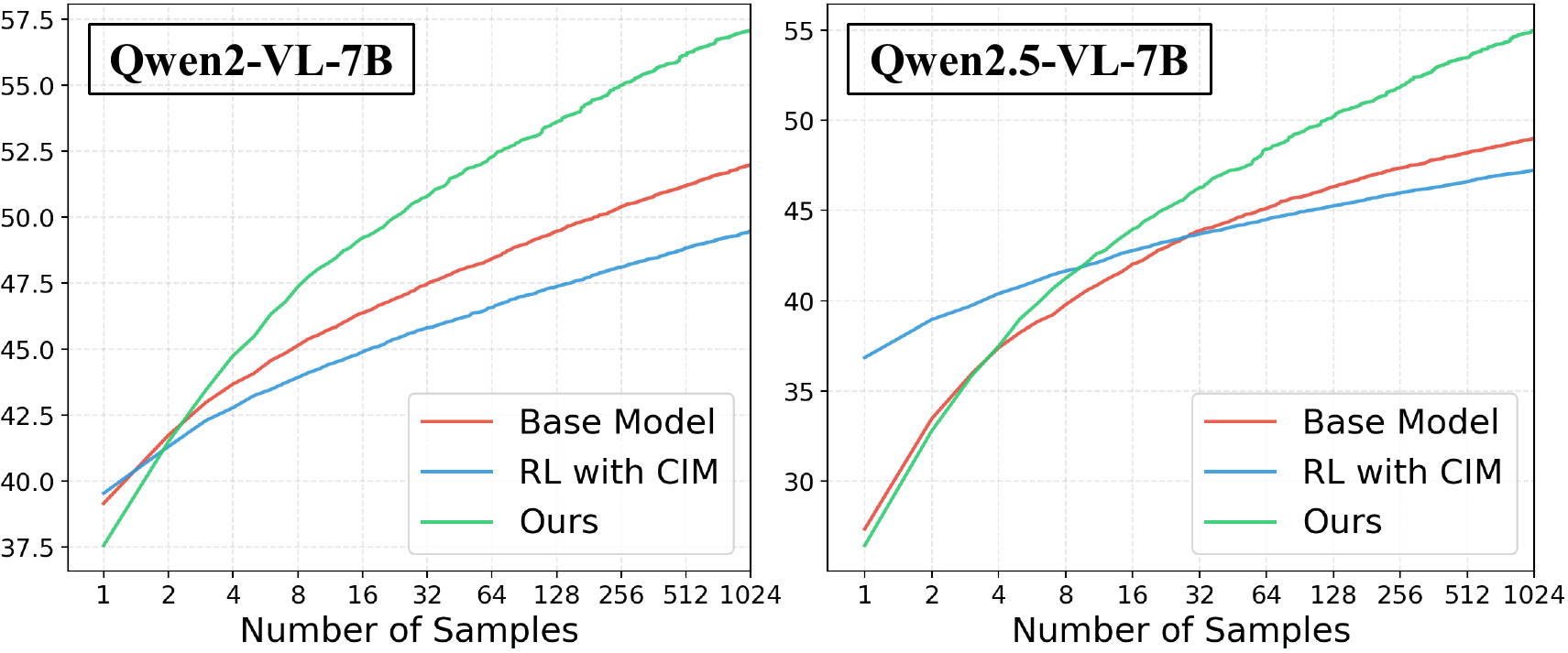}
  \caption{\textbf{Capability boundary analysis with max@$k$.}}
  \label{fig:max@k}
\end{figure}

\begin{table*}[!ht]
\tablestyle{3pt}{1.25}
\centering
\begin{tabular}{lllcccccc}
\hline
\multirow{2}{*}{\makecell{Base \\Model}} &
\multirow{2}{*}{Reward} &
\multirow{2}{*}{Method} &
\multicolumn{3}{c}{COCO-LN500} &
\multicolumn{3}{c}{DOCCI500} \\
\cline{4-9}
&&&
\makecell{Objects\\F1} &
\makecell{Attributes\\F1} &
\makecell{Relations\\QA} &
\makecell{Objects\\F1} &
\makecell{Attributes\\F1} &
\makecell{Relations\\QA} \\
\hline

\multirow{8}{*}{\makecell{LLaVA1.5\\-7B}}
& \multirow{2}{*}{--} & \vb Base
& 67.56 & 42.34 & 14.38
& 59.41 & 48.01 & 9.19 \\
& & \vb SFT
& 73.45 & 54.25 & 28.59
& 68.44 & 53.93 & 19.87 \\
\cline{2-9}

& \multirow{2}{*}{CLIP} & \vb GRPO
& 66.66 & 53.15 & 22.58
& 61.77 & 53.62 & 18.42 \\
& & \oursbg \cb Ours
& \oursbg \textbf{72.03} ($\uparrow$5.4)
& \oursbg \textbf{54.73} ($\uparrow$1.6)
& \oursbg \textbf{30.38} ($\uparrow$7.8)
& \oursbg \textbf{68.51} ($\uparrow$6.7)
& \oursbg \textbf{57.68} ($\uparrow$4.1)
& \oursbg \textbf{23.86} ($\uparrow$5.4) \\
\cline{2-9}

& \multirow{2}{*}{SC} & \vb GRPO
& 69.49 & 50.44 & 21.77
& 62.52 & 53.80 & 17.86 \\
& & \oursbg \cb Ours
& \oursbg \textbf{74.08} ($\uparrow$4.6)
& \oursbg \textbf{54.57} ($\uparrow$4.1)
& \oursbg \textbf{35.17} ($\uparrow$13.4)
& \oursbg \textbf{70.25} ($\uparrow$7.7)
& \oursbg \textbf{56.21} ($\uparrow$2.4)
& \oursbg \textbf{26.92} ($\uparrow$9.1) \\
\cline{2-9}

& \multirow{2}{*}{CIM} & \vb GRPO
& 69.80 & 54.38 & 24.98
& 63.38 & \textbf{56.28} & 19.87 \\
& & \oursbg \cb Ours
& \oursbg \textbf{74.74} ($\uparrow$4.9)
& \oursbg \textbf{54.73} ($\uparrow$0.4)
& \oursbg \textbf{34.93} ($\uparrow$10.0)
& \oursbg \textbf{69.49} ($\uparrow$6.1)
& \oursbg 56.13 ($\downarrow$0.2)
& \oursbg \textbf{27.93} ($\uparrow$8.1) \\
\hline

\multirow{8}{*}{\makecell{Qwen2\\-VL-7B}}
& \multirow{2}{*}{--} & \vb Base
& 69.47 & 48.68 & 20.47
& 66.47 & 52.65 & 17.57 \\
& & \vb SFT
& 75.37 & 56.54 & 36.39
& 69.50 & 55.50 & 27.65 \\
\cline{2-9}

& \multirow{2}{*}{CLIP} & \vb GRPO
& 69.04 & 53.37 & 26.48
& 67.30 & 54.63 & 26.28 \\
& & \oursbg \cb Ours
& \oursbg \textbf{75.33} ($\uparrow$6.3)
& \oursbg \textbf{58.35} ($\uparrow$5.0)
& \oursbg \textbf{35.26} ($\uparrow$8.8)
& \oursbg \textbf{71.88} ($\uparrow$4.6)
& \oursbg \textbf{58.37} ($\uparrow$3.7)
& \oursbg \textbf{32.73} ($\uparrow$6.5) \\
\cline{2-9}

& \multirow{2}{*}{SC} & \vb GRPO
& 76.80 & 57.49 & 30.46
& 72.49 & 57.75 & 23.50 \\
& & \oursbg \cb Ours
& \oursbg \textbf{77.77} ($\uparrow$1.0)
& \oursbg \textbf{58.79} ($\uparrow$1.3)
& \oursbg \textbf{44.60} ($\uparrow$14.1)
& \oursbg \textbf{73.29} ($\uparrow$0.8)
& \oursbg \textbf{58.88} ($\uparrow$1.1)
& \oursbg \textbf{36.40} ($\uparrow$12.9) \\
\cline{2-9}

& \multirow{2}{*}{CIM} & \vb GRPO
& 75.80 & 58.22 & 38.71
& 71.43 & \textbf{59.18} & 32.12 \\
& & \oursbg \cb Ours
& \oursbg \textbf{78.18} ($\uparrow$2.4)
& \oursbg \textbf{59.28} ($\uparrow$1.1)
& \oursbg \textbf{44.19} ($\uparrow$5.5)
& \oursbg \textbf{73.51} ($\uparrow$2.1)
& \oursbg 58.99 ($\downarrow$0.2)
& \oursbg \textbf{40.15} ($\uparrow$8.0) \\
\hline

\multirow{8}{*}{\makecell{Qwen2.5\\-VL-7B}}
& \multirow{2}{*}{--} & \vb Base
& 65.37 & 46.25 & 23.76
& 65.06 & 52.27 & 24.35 \\
& & \vb SFT
& 75.72 & 57.09 & 39.64
& 71.94 & 58.28 & 34.38 \\
\cline{2-9}

& \multirow{2}{*}{CLIP} & \vb GRPO
& 68.32 & \textbf{53.90} & 23.80
& 68.21 & 55.32 & 24.79 \\
& & \oursbg \cb Ours
& \oursbg \textbf{70.52} ($\uparrow$2.2)
& \oursbg 53.80 ($\downarrow$0.1)
& \oursbg \textbf{30.26} ($\uparrow$6.5)
& \oursbg \textbf{69.39} ($\uparrow$1.2)
& \oursbg \textbf{55.70} ($\uparrow$0.4)
& \oursbg \textbf{30.27} ($\uparrow$5.5) \\
\cline{2-9}

& \multirow{2}{*}{SC} & \vb GRPO
& \textbf{77.52} & 56.71 & 31.19
& \textbf{72.81} & 57.93 & 28.01 \\
& & \oursbg \cb Ours
& \oursbg 76.85 ($\downarrow$0.7)
& \oursbg \textbf{58.96} ($\uparrow$2.3)
& \oursbg \textbf{41.06} ($\uparrow$9.9)
& \oursbg \textbf{72.81} ($\uparrow$0.0)
& \oursbg \textbf{58.51} ($\uparrow$0.6)
& \oursbg \textbf{36.80} ($\uparrow$8.8) \\
\cline{2-9}

& \multirow{2}{*}{CIM} & \vb GRPO
& 77.59 & 58.51 & 44.15
& 71.88 & 59.08 & 34.70 \\
& & \oursbg \cb Ours
& \oursbg \textbf{77.80} ($\uparrow$0.2)
& \oursbg \textbf{59.26} ($\uparrow$0.8)
& \oursbg \textbf{46.02} ($\uparrow$1.9)
& \oursbg \textbf{72.80} ($\uparrow$0.9)
& \oursbg \textbf{59.77} ($\uparrow$0.7)
& \oursbg \textbf{38.65} ($\uparrow$4.0) \\
\hline
\end{tabular}
\caption{
\textbf{Performance comparison of Reinforcement Learning with GRPO and Re$^3$Cap across multiple reward functions on COCO-LN500~\cite{pont2020connecting} and DOCCI500~\cite{onoe2024docci}.}
Constrained by the reasoning capacity of base models, GRPO struggles to surpass the performance of task-specific SFT models, especially on weaker LVLMs such as LLaVA-1.5-7B~\cite{liu2023improvedllava}.
In contrast, our method introduces a novel reasoning strategy to generate unexplored caption candidates, consistently improving image captioning performance across weaker and stronger base models, including Qwen2-VL-7B~\cite{wang2024qwen2} and Qwen2.5-VL-7B~\cite{bai2025qwen2}.
Moreover, using a more accurate reward further improves the performance of ours.
}
\label{exp:grpo}
\end{table*}

\subsection{Reasoning Capability Analysis}
Inspired by \citet{yue2025does}, we further evaluate whether our retrieval-guided reasoning strategy enables the LVLM to explore caption candidates beyond those already covered by the base model. Specifically, for each image, we sample multiple candidate captions from each model and evaluate them on COCO-LN500~\cite{pont2020connecting} using the BLEU-4~\cite{papineni2002bleu} score. Since BLEU-4 is a continuous metric, we adopt max@$k$, a continuous generalization of pass@$k$~\cite{bagirov2025best}, to measure the best achievable caption quality under a large sampling budget. 
We compute max@$k$ using the unbiased low-variance estimator proposed by \citet{walder2026pass}.

As shown in ~\Cref{fig:max@k}, the reinforcement learning method using CIM~\cite{jia2026cim} as the reward function achieves strong performance at $k=1$, indicating that conventional RL effectively improves sampling efficiency. However, as $k$ increases, it grows more slowly and eventually falls below the base model, suggesting that it tends to narrow the output distribution and does not preserve sufficient exploration diversity. In contrast, our retrieval-guided reasoning strategy, without any training, consistently benefits from larger sampling budgets and surpasses the base model as $k$ increases. This trend indicates that our method encourages the model to generate more diverse and previously unexplored caption candidates. Therefore, the results demonstrate that our method can expand the capability boundary of the base model. The experiments in~\Cref{sec:4.3} further validate this conclusion by showing that, when incorporated into reinforcement learning, our strategy brings consistent improvements across different LVLMs, benchmarks, and reward functions.

\subsection{Reinforcement Learning on Base Model}
\label{sec:4.3}
We verify the generalization of our method via various LVLMs and evaluate performance on benchmarks~\cite{pont2020connecting,onoe2024docci}. 
In tables, Base means the LVLMs without any task-specific training, and SFT denotes the LVLMs supervised fine-tuned on the RefinedCaps dataset~\cite{zhang2025sc}. 
GRPO and Ours denote models trained from the base model with GRPO and Re$^3$Cap, respectively.

As shown in \Cref{exp:grpo}, the results demonstrate that our method consistently outperforms GRPO across multiple LVLMs and with various reward functions, especially on the more challenging relation reasoning task.
For instance, on the QA score of the Relations evaluation, our method achieves improvements by 8.64\% on COCO-LN500~\cite{pont2020connecting} and 7.57\% on DOCCI500~\cite{onoe2024docci}, averaged across multiple base LVLMs and reward functions.
Moreover, the gains are particularly pronounced when using weaker base LVLMs and reward functions.
With CLIP~\cite{cho2022fine} as the reward function, our method achieves gains of 4.39\% in Objects F1, 2.44\% in Attributes F1, and 6.74\% in Relations QA, averaged across multiple base LVLMs and both benchmarks.
For LLaVA1.5-7B~\cite{liu2023improvedllava} as the base LVLM, our method yields gains of 5.91\% in Objects F1, 2.06\% in Attributes F1, and 8.95\% in Relations QA, averaged across multiple reward functions and both benchmarks.
Notably, using LLaVA1.5-7B~\cite{liu2023improvedllava} as the base LVLM, our method even outperforms SFT across multiple reward functions and both benchmarks, whereas GRPO underperforms SFT.
This demonstrates that GRPO is constrained by the reasoning capacity of base models, making it difficult to surpass the performance of models supervised fine-tuned on task-specific datasets.
In contrast, our method can generate previously unexplored caption candidates from base LVLMs by injecting a novel reasoning strategy, leading to substantial gains.
For Qwen2-VL-7B~\cite{wang2024qwen2} and Qwen2.5-VL-7B~\cite{bai2025qwen2} as the base LVLMs, GRPO can surpass SFT when using CIM~\cite{jia2026cim} as the reward function. 
Even on these strong base LVLMs, our method further improves performance, indicating its effectiveness.
The above results demonstrate that Re$^3$Cap significantly enhances the image captioning capability of LVLMs with distinct architectures.

\begin{table*}[!ht]
\tablestyle{3pt}{1.25}
\centering
\begin{tabular}{lllcccccc}
\hline
\multirow{2}{*}{\makecell{Base \\Model}} &
\multirow{2}{*}{Reward} &
\multirow{2}{*}{Method} &
\multicolumn{3}{c}{COCO-LN500} &
\multicolumn{3}{c}{DOCCI500} \\
\cline{4-9}
&&&
\makecell{Objects\\F1} &
\makecell{Attributes\\F1} &
\makecell{Relations\\QA} &
\makecell{Objects\\F1} &
\makecell{Attributes\\F1} &
\makecell{Relations\\QA} \\
\hline

\multirow{6}{*}{\makecell{Qwen2\\-VL-7B}}
& \multirow{2}{*}{--}
& \vb VCD
& 69.85 & 48.71 & 26.77
& 67.57 & 53.80 & 25.03 \\
& & \vb INTER
& 69.81 & 48.58 & 26.93
& 67.42 & 53.70 & 24.67 \\
\cline{2-9}

& \multirow{2}{*}{SC}
& \vb SC-Captioner
& 76.37 & 57.56 & 38.51
& 71.63 & 57.67 & 30.51 \\
& & \oursbg \cb Ours
& \oursbg \textbf{77.77} ($\uparrow$1.4)
& \oursbg \textbf{58.79} ($\uparrow$1.2)
& \oursbg \textbf{44.60} ($\uparrow$6.1)
& \oursbg \textbf{73.29} ($\uparrow$1.7)
& \oursbg \textbf{58.88} ($\uparrow$1.2)
& \oursbg \textbf{36.40} ($\uparrow$5.9) \\
\cline{2-9}

& \multirow{2}{*}{CIM}
& \vb SFT+CIM
& 76.65 & 58.09 & 42.12
& 73.87 & 58.68 & 36.32 \\
& & \oursbg \cb Ours
& \oursbg \textbf{78.18} ($\uparrow$1.5)
& \oursbg \textbf{59.28} ($\uparrow$1.2)
& \oursbg \textbf{44.19} ($\uparrow$2.1)
& \oursbg \textbf{73.92} ($\uparrow$0.1)
& \oursbg \textbf{58.99} ($\uparrow$0.3)
& \oursbg \textbf{40.15} ($\uparrow$3.8) \\
\hline
\end{tabular}
\caption{
\textbf{Performance comparison with state-of-the-art methods on Qwen2-VL-7B~\cite{wang2024qwen2} over COCO-LN500~\cite{pont2020connecting} and DOCCI500~\cite{onoe2024docci}.}
The results demonstrate the superiority of our proposed reinforcement learning framework when compared with state-of-the-art methods.
}
\label{exp:sota}
\end{table*}

\subsection{SOTA Comparison}
To further verify the superiority of our method, we conduct experiments to compare it with state-of-the-art methods.
As shown in~\Cref{exp:sota}, VCD~\cite{leng2024mitigating} and INTER~\cite{dong2025inter} are training-free methods designed to mitigate hallucinations in LVLMs.
SC-Captioner~\cite{zhang2025sc} and SFT+CIM~\cite{jia2026cim} both perform reinforcement learning with their respective reward functions after first applying Supervised Fine-Tuning (SFT) to the base LVLMs.
In contrast, Ours refers to the base model solely optimized by Re$^3$Cap via reinforcement learning, without supervised fine-tuning on task-specific datasets.

The results demonstrate that our approach achieves superior performance in image captioning, even compared with SFT-based methods.
Specifically, on the more challenging relation reasoning task, our method improves Relations QA by 4.08\% on COCO-LN500~\cite{pont2020connecting} and 4.86\% on DOCCI500~\cite{onoe2024docci}, averaged across multiple reward functions.
Additionally, our method achieves gains of 1.47\% in Objects F1 and 1.21\% in Attributes F1 on COCO-LN500~\cite{pont2020connecting}, averaged across multiple reward functions.
Notably, our method achieves these improvements with only a single-stage RL training, whereas SC-Captioner and SFT+CIM rely on a two-stage pipeline (SFT followed by RL). 
Moreover, our method outperforms the training-free methods across all metrics by a large margin.
Such results indicate that by injecting the reasoning strategy during reinforcement learning, our method achieves superior performance in image captioning. 

\subsection{Ablation Study}
In this section, we present an ablation study to quantitatively evaluate the effectiveness of each core component (CRS and CQA) within our framework.
As shown in \Cref{exp:ablation}, the first row denotes the model trained solely with GRPO~\cite{shao2024deepseekmath}.
The middle two rows refer to models optimized by reinforcement learning that use CRS and CQA as their reasoning strategy, respectively. 
The last row denotes the model trained using Re$^3$Cap, combined with CRS and CQA.

The results in \Cref{exp:ablation} show that each core component achieves consistent performance improvements.
Specifically, CRS achieves improvements of 1.03\% in Objects F1, 0.76\% in Attributes F1, and 2.19\% in Relations QA on COCO-LN500.
Moreover, CQA achieves improvements of 1.34\% in Objects F1, 0.82\% in Attributes F1, and 3.98\% in Relations QA.
Such results indicate CRS improves performance by guiding the LVLM to retain accurate descriptions, and CQA guides the LVLM to mitigate hallucinations and reduce omissions, thereby further improving performance.
Most importantly, our method achieves the best performance by combining the complementary CRS and CQA.

\begin{table}[!t]
\tablestyle{5pt}{1.25}
\centering
\begin{tabular}{ccccc}
\hline
\multicolumn{2}{c}{Components} &
Objects &
Attributes &
Relations \\
\cline{1-5}
CRS & CQA & F1 & F1 & QA \\
\hline

\xmark & \xmark
& 75.80 & 58.22 & 38.71 \\
\cmark & \xmark
& 76.83 & 58.98 & 40.90 \\
\xmark & \cmark
& 77.14 & 59.04 & 42.69 \\
\oursbg \cmark & \oursbg \cmark
& \oursbg \textbf{78.18}
& \oursbg \textbf{59.28}
& \oursbg \textbf{44.19} \\
\hline
\end{tabular}
\caption{
\textbf{Ablation study of core components on COCO-LN500~\cite{pont2020connecting} using Qwen2-VL-7B~\cite{wang2024qwen2} with CIM~\cite{jia2026cim} as the reward function.}
}
\label{exp:ablation}
\end{table}

\section{Conclusion}
In this paper, we present Re$^3$Cap, a reinforcement learning framework that consistently outperforms previous Supervised Fine-Tuning (SFT) approaches in image captioning. 
Our key observation is that discrepancies between retrieval results reveal potential hallucinations and omitted visual details in image captions, providing an informative signal for assessing caption quality.
Building on this insight, we further propose a retrieval-based reasoning strategy that guides LVLMs to generate previously unexplored caption candidates.
By performing reinforcement learning on these newly explored candidates, the model effectively expands the caption space and refines its generation behavior.
Extensive experiments demonstrate that our proposed Re$^3$Cap enables LVLMs to achieve consistently superior performance in image captioning, even compared with strong Supervised Fine-Tuning (SFT) baselines. 
Overall, this work enhances the reasoning capabilities of LVLMs in reinforcement learning and offers a new perspective on improving model performance in image captioning. 
We hope the proposed framework provides more insights for future research in both multimodal reasoning and caption generation.

\section*{Limitations}
Specifically, the effectiveness of our method depends on the quality and scale of the retrieval set. 
When the dataset is too small, many image captions fail to retrieve relevant results. 
In such cases, our method may degenerate into a simple reinforcement learning approach.
We believe that increasing the scale and diversity of the retrieval set would improve the robustness of our approach.
\section*{Ethical Considerations}
This work aims to improve image captioning by introducing a retrieval-guided refinement strategy during reinforcement learning. All experiments are conducted on publicly available image-caption datasets and benchmarks. As in prior work, these datasets and evaluation protocols may contain social biases, annotation artifacts, sampling biases, or other imperfections that can affect model behavior and evaluation outcomes. Beyond the risks already associated with multimodal model training, retrieval, and evaluation on existing public datasets, we do not identify additional ethical risks introduced specifically by our method.

\bibliography{custom}

\clearpage
\appendix

\section{Prompt Templates}
We use a concise prompt to generate the initial caption. For Re$^3$Cap, the model analyzes the initial caption with the reasoning strategy and produces guidance, which is then injected into the response to prompt the model to generate a refined caption. For Relation QA, we prompt Qwen3~\cite{yang2025qwen3} models to answer the given questions based on the candidate captions. The detailed prompts are shown in Figure \ref{fig:prompt}.

\begin{figure*}[h]
  \centering
    \begin{promptbox}{Prompt for Initial Caption}
    \textbf{User:} Caption this image as accurately as possible, without speculation. Describe what you see.
    
    \textbf{Assistant:}
    \end{promptbox}
    \begin{promptbox}{Prompt for Re$^3$Cap}
    \textbf{User:} Caption this image as accurately as possible, without speculation. Describe what you see.
    
    \textbf{Assistant:} <Initial Caption>
    
    Wait. Let me check this caption again against the image.
    
    I should KEEP: <Correctness>
    
    I should ADD: <Omissions>
    
    I should REMOVE: <Hallucinations>

    Final caption:
    \end{promptbox}
    \begin{promptbox}{Prompt for Relation QA}
    \textbf{User:} I will give you a passage of caption. Please answer the following 5 questions with "Yes", "No", or "n/a" based on the given caption. Output like this: 1: Yes, 2: No, 3: Yes, 4: n/a, 5: Yes. Don't output extra text.

    Caption: <Caption>
    
    Questions:1.<Question1> 2.<Question2> 3.<Question3> 4.<Question4> 5.<Question5>
    
    \textbf{Assistant:}
    \end{promptbox}
  \caption{Prompt example for initial caption, Re$^3$Cap, and relation evaluation.}
  \label{fig:prompt}
\end{figure*}

\section{Additional Experiments}\label{sec:b}
\subsection{Reinforcement Learning on More LVLMs}\label{sec:b.1}
\begin{table*}[!ht]
\tablestyle{3pt}{1.25}
\centering
\begin{tabular}{lllcccccc}
\hline
\multirow{2}{*}{\makecell{Base \\Model}} &
\multirow{2}{*}{Reward} &
\multirow{2}{*}{Method} &
\multicolumn{3}{c}{COCO-LN500} &
\multicolumn{3}{c}{DOCCI500} \\
\cline{4-9}
&&&
\makecell{Objects\\F1} &
\makecell{Attributes\\F1} &
\makecell{Relations\\QA} &
\makecell{Objects\\F1} &
\makecell{Attributes\\F1} &
\makecell{Relations\\QA} \\
\hline

\multirow{8}{*}{\makecell{InternVL3\\-8B}}
& \multirow{2}{*}{--}
& \vb Base
& 71.00 & 50.66 & 26.44
& 66.08 & 53.72 & 25.11 \\
& & \vb SFT
& 76.42 & 57.79 & 42.32
& 72.72 & 58.59 & 35.31 \\
\cline{2-9}

& \multirow{2}{*}{CLIP}
& \vb GRPO
& 72.14 & \textbf{54.84} & 30.83
& 68.70 & \textbf{57.58} & 28.13 \\
& & \oursbg \cb Ours
& \oursbg \textbf{74.79} ($\uparrow$2.7)
& \oursbg 54.42 ($\downarrow$0.4)
& \oursbg \textbf{35.46} ($\uparrow$4.6)
& \oursbg \textbf{71.15} ($\uparrow$2.5)
& \oursbg 54.85 ($\downarrow$2.7)
& \oursbg \textbf{28.81} ($\uparrow$0.7) \\
\cline{2-9}

& \multirow{2}{*}{SC}
& \vb GRPO
& 75.77 & 57.14 & 35.58
& 70.72 & 57.78 & 28.38 \\
& & \oursbg \cb Ours
& \oursbg \textbf{78.51} ($\uparrow$2.7)
& \oursbg \textbf{59.50} ($\uparrow$2.4)
& \oursbg \textbf{45.21} ($\uparrow$9.6)
& \oursbg \textbf{74.11} ($\uparrow$3.4)
& \oursbg \textbf{58.19} ($\uparrow$0.4)
& \oursbg \textbf{36.84} ($\uparrow$8.5) \\
\cline{2-9}

& \multirow{2}{*}{CIM}
& \vb GRPO
& 76.14 & 58.70 & 38.67
& 70.47 & 59.26 & 30.39 \\
& & \oursbg \cb Ours
& \oursbg \textbf{78.68} ($\uparrow$2.5)
& \oursbg \textbf{58.99} ($\uparrow$0.3)
& \oursbg \textbf{44.19} ($\uparrow$5.5)
& \oursbg \textbf{73.34} ($\uparrow$2.9)
& \oursbg \textbf{59.78} ($\uparrow$0.5)
& \oursbg \textbf{36.60} ($\uparrow$6.2) \\
\hline

\multirow{8}{*}{\makecell{Qwen3\\-VL-8B}}
& \multirow{2}{*}{--}
& \vb Base
& 72.64 & 53.73 & 37.57
& 72.68 & 56.08 & 40.67 \\
& & \vb SFT
& 76.59 & 57.14 & 42.24
& 72.51 & 57.37 & 38.65 \\
\cline{2-9}

& \multirow{2}{*}{CLIP}
& \vb GRPO
& 73.91 & 54.72 & 39.84
& 72.93 & 56.31 & 39.82 \\
& & \oursbg \cb Ours
& \oursbg \textbf{74.86} ($\uparrow$1.0)
& \oursbg \textbf{55.83} ($\uparrow$1.1)
& \oursbg \textbf{41.37} ($\uparrow$1.5)
& \oursbg \textbf{73.24} ($\uparrow$0.3)
& \oursbg \textbf{56.73} ($\uparrow$0.4)
& \oursbg \textbf{41.36} ($\uparrow$1.5) \\
\cline{2-9}

& \multirow{2}{*}{SC}
& \vb GRPO
& 75.11 & 54.98 & 42.85
& 72.96 & 56.64 & 40.67 \\
& & \oursbg \cb Ours
& \oursbg \textbf{76.45} ($\uparrow$1.3)
& \oursbg \textbf{57.31} ($\uparrow$2.3)
& \oursbg \textbf{46.26} ($\uparrow$3.4)
& \oursbg \textbf{73.76} ($\uparrow$0.8)
& \oursbg \textbf{57.02} ($\uparrow$0.4)
& \oursbg \textbf{41.92} ($\uparrow$1.3) \\
\cline{2-9}

& \multirow{2}{*}{CIM}
& \vb GRPO
& 75.21 & 56.03 & 39.52
& 72.27 & 57.39 & 38.17 \\
& & \oursbg \cb Ours
& \oursbg \textbf{76.92} ($\uparrow$1.7)
& \oursbg \textbf{57.98} ($\uparrow$2.0)
& \oursbg \textbf{45.86} ($\uparrow$6.3)
& \oursbg \textbf{73.60} ($\uparrow$1.3)
& \oursbg \textbf{57.94} ($\uparrow$0.6)
& \oursbg \textbf{42.64} ($\uparrow$4.5) \\
\hline
\end{tabular}
\caption{\textbf{Performance comparison of reinforcement learning with GRPO and Re$^3$Cap on more LVLMs.} We evaluate InternVL3-8B~\cite{zhu2025internvl3} and Qwen3-VL-8B~\cite{bai2025qwen3} with different reward functions on COCO-LN500~\cite{pont2020connecting} and DOCCI500~\cite{onoe2024docci}. Re$^3$Cap consistently improves over GRPO across most metrics, especially on the Relations QA task.}
\label{exp:grpo2}
\end{table*}
We further adopt InternVL3-8B~\cite{zhu2025internvl3} and Qwen3-VL-8B~\cite{bai2025qwen3} as the base LVLM to validate the effectiveness of our method. As shown in \Cref{exp:grpo2}, the results demonstrate that our method consistently outperforms GRPO across multiple reward functions, especially on the more challenging relation reasoning task.
Specifically, for InternVL3-8B, Re$^3$Cap improves the Relations QA score over GRPO by an average of 6.59\% on COCO-LN500~\cite{pont2020connecting} and 5.12\% on DOCCI500~\cite{onoe2024docci} across multiple reward functions. 
For Qwen3-VL-8B, Re$^3$Cap also brings consistent improvements, achieving average gains of 3.76\% on COCO-LN500 and 2.42\% on DOCCI500 in Relations QA over GRPO.
Moreover, GRPO fails to surpass SFT under the CIM~\cite{jia2026cim} and SC~\cite{zhang2025sc} reward functions, whereas our method consistently outperforms SFT under these reward signals.
The above results demonstrate that our proposed Re$^3$Cap significantly enhances the image captioning capability of LVLMs.

\subsection{Robustness of the Choice of Hyperparameters}
\begin{figure*}[t]
  \centering
    \includegraphics[width=\linewidth]{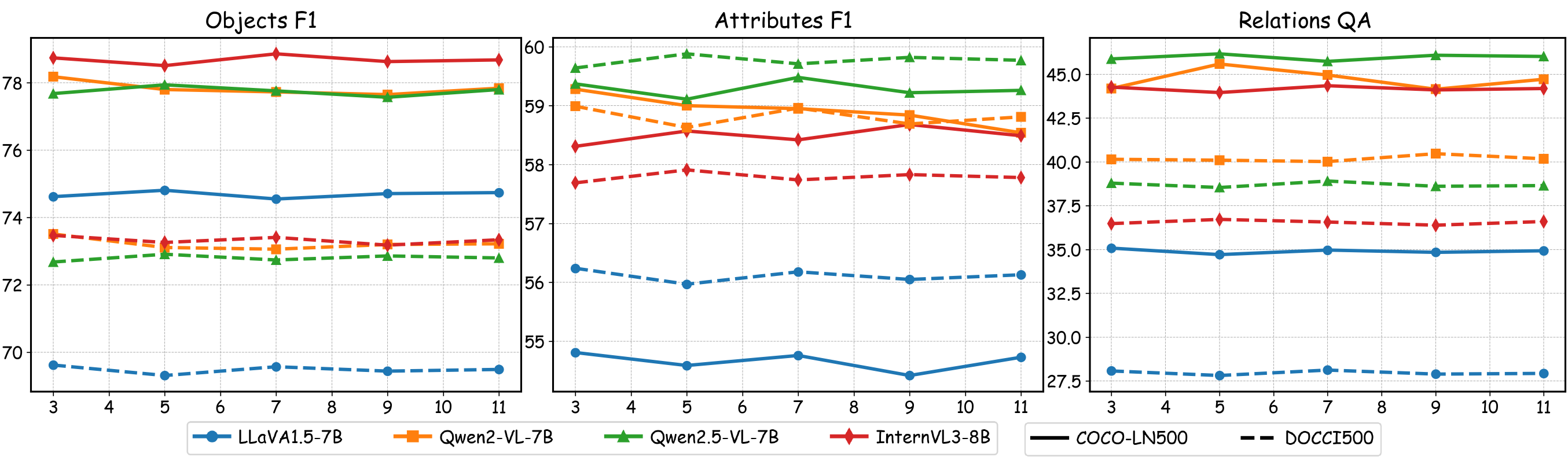}
    \caption{\textbf{Robustness study of the retrieval number $K$.} We evaluate Re$^3$Cap with different retrieval numbers $K$ on COCO-LN500~\cite{pont2020connecting} and DOCCI500~\cite{onoe2024docci} across multiple LVLMs. The performance remains stable across different values of $K$, demonstrating that our method is robust to the choice of retrieval number.}
  \label{fig:robustness_K}
\end{figure*}

\begin{figure*}[t]
  \centering
    \includegraphics[width=\linewidth]{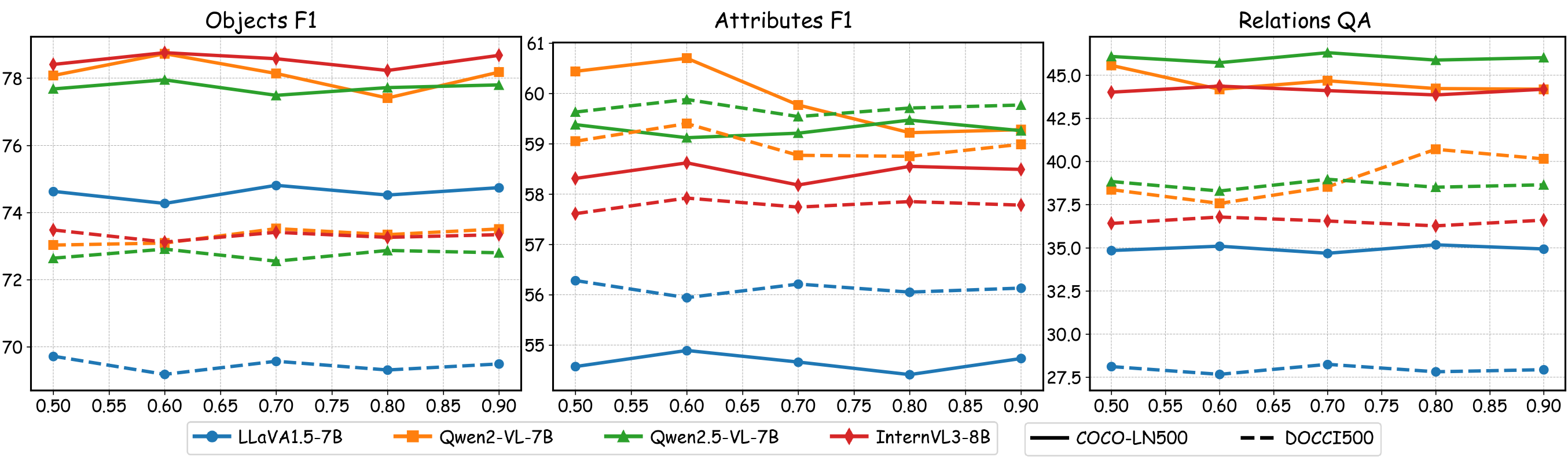}
  \caption{\textbf{Robustness study of the similarity threshold $\tau$.} We evaluate Re$^3$Cap with different similarity thresholds $\tau$ for graph edge construction on COCO-LN500~\cite{pont2020connecting} and DOCCI500~\cite{onoe2024docci} across multiple LVLMs. The results show only minor variations across different thresholds, indicating that our method is robust to the choice of graph construction threshold.}
  \label{fig:robustness_tau}
\end{figure*}

We conduct experiments to evaluate the effects of two hyperparameters on model performance: the retrieval number $K$ and the similarity threshold $\tau$ used for edge construction. As shown in~\Cref{fig:robustness_K} and~\Cref{fig:robustness_tau}, we vary $K$ from 3 to 11 and $\tau$ from 0.5 to 0.9.
The results show only minor performance variations across different settings. Specifically, on COCO-LN500~\cite{pont2020connecting} with Qwen2-VL-7B~\cite{wang2024qwen2}, the variations in Objects F1, Attributes F1, and Relation QA are within 0.53\%, 0.74\%, and 1.44\%, respectively, across different $K$ values, and within 1.32\%, 1.48\%, and 1.38\%, respectively, across different $\tau$ values. Overall, such results demonstrate that our method is robust to the choice of hyperparameters.

\subsection{Robustness across Diverse Encoders}
We conduct experiments to evaluate the impact of different encoders used in our method for retrieval and graph edge construction.
As shown in~\Cref{exp:robustness}, we use either DINOv3 ViT-L/16~\cite{simeoni2025dinov3} or OpenCLIP ViT-H/14~\cite{cherti2023reproducible} as the image encoder, and SBERT~\cite{reimers2019sentence} with a MiniLM-base~\cite{wang2020minilm} or MPNet-base~\cite{song2020mpnet} backbone as the text encoder.
The results show only minor performance variations across different encoders.
Specifically, the variations in Objects F1, Attributes F1, and Relation QA are within 0.46\%, 0.79\%, and 0.94\%, respectively, indicating strong robustness to the choice of both image and text encoders.
Overall, such results demonstrate that our method is robust to the potential information loss and representation discrepancies introduced by different encoders.

\begin{table*}[!t]
\tablestyle{3pt}{1.25}
\centering
\begin{tabular}{llccccccc}
\hline
\multirow{2}{*}{\makecell{Image \\Encoder}} & 
\multirow{2}{*}{\makecell{Text \\Encoder}} & 
\multicolumn{3}{c}{Objects} &
\multicolumn{3}{c}{Attributes} &
Relations \\
\cline{3-9}
 &  & Precision & Recall & F1 & Precision & Recall & F1 & QA \\
\hline

\multirow{2}{*}{\makecell{DINOv3 \\ViT-L/16}} & MiniLM & 79.24 & 77.23 & 77.77 & 70.48 & 55.89 & 59.03 & 45.13 \\

& \oursbg MPNet & \oursbg 80.60 & \oursbg 76.88 & \oursbg 78.23 & \oursbg 71.34 & \oursbg 54.82 & \oursbg 58.57 & \oursbg 44.68 \\

\multirow{2}{*}{\makecell{OpenCLIP \\ViT-H/14}}  & MiniLM & 79.77 & 77.03 & 77.91 & 69.99 & 55.29 & 58.49 & 45.04 \\

& \oursbg MPNet & \oursbg 79.62 & \oursbg 77.70 & \oursbg 78.18 & \oursbg 71.00 & \oursbg 56.03 & \oursbg 59.28 & \oursbg 44.19 \\

\hline
\end{tabular}
\caption{
\textbf{Robustness study of diverse encoders on COCO-LN500~\cite{pont2020connecting} using Qwen2-VL-7B~\cite{wang2024qwen2} with  CIM~\cite{jia2026cim} as the reward function.}
Our method maintains stable performance in image captioning when leveraging various encoders as the retrieval model.
The experimental results indicate that our proposed Re$^3$Cap is robust to diverse encoders.
}
\label{exp:robustness}
\end{table*}

\subsection{Analysis of Computational Overhead}
Taking SC~\cite{zhang2025sc} as the reward function as an example, GRPO requires 192 GPU hours on the NVIDIA A100 to converge, whereas our method converges in approximately 216 GPU hours under the same experimental settings. This corresponds to only an additional 24 GPU hours, or about 12.5\% more training time, while achieving significant performance gains. Moreover, the memory overhead of our method is comparable to that of GRPO.

\end{document}